%% file: report.tex
\documentclass[11pt]{article}
\usepackage{acl2015}
\usepackage{times}
\usepackage{url}
\usepackage{latexsym}

\usepackage[utf8]{inputenc}
\usepackage{algorithm}
\usepackage{algpseudocode}
\usepackage{booktabs}
\usepackage{fullpage}
\usepackage[utf8]{inputenc}
\usepackage{hyperref}
\usepackage{amsbsy}

\usepackage{amsfonts}
\usepackage{amsmath}
\usepackage{amssymb}
\usepackage{booktabs}

\usepackage{tikz}
\usetikzlibrary{fit}
\usetikzlibrary{positioning}
\usepackage{array,multirow,graphicx}
\usepackage{balance}
\title{e-Commerce product classification: our participation at cDiscount 2015 challenge}

\author{Ioannis Partalas \\
  Viseo R\&D, France  \\
  {\tt ioannis.partalas@viseo.com} \\\And
  Georgios Balikas \\
  University of Grenoble Alpes, France  \\
  {\tt georgios.balikas@imag.fr} \\}

\date{}

\begin{document}

\maketitle

\begin{abstract}
This report describes  our participation in the cDiscount 2015 challenge where the goal was to classify product items in a predefined
taxonomy of products.
Our best submission yielded an accuracy score of 64.20\% in the private part of the leaderboard and we were ranked 10th out of 175 participating teams. 
We followed a text classification approach employing mainly linear models. The final solution was a weighted voting system which combined
a variety of trained models.
\end{abstract}

\input{introduction}

\input{product_classification}

\input{our_approach}

\input{results}

\section*{Acknowledgements}
We would like to thank the AMA team of the University of Grenoble-Alpes for providing us the machines where we ran our algorithms.

\end{document}

%% file: introduction.tex
\section{Introduction}
%Classification concerns with the prediction of categories a new observation belongs, on the basis of a training set of data containing observations (or instances) whose category membership is known is called classification and is a well-studied problem in the area of Machine Learning. When the observations to be classified are text spans the problem is often referred to as text classification. Over the past years, several challenges have been organised in order to assess the performance of the state-of-the-art methods that perform classification. We cite for example the BioASQ \cite{tsatsaronis2015overview,balikas2014results} and LSHTC \cite{partalas2015lshtc} challenges that were organised by academic consurtiums and the WISE and the XXX challenges that were run in Kaggle\footnote{\texttt{www.kaggle.com}} and in datascience.net respectively that were organised by companies to address industrial problems. 
In this report we present our participation in the cDicount 2015 product classification challenge that was organised in the \texttt{www.datascience.net} platform.
The organisers provided a large collection of product items containing mainly its textual description. We followed a text classification
approach using mainly linear models (like SVMs) as our base models. Our final solution consisted of a weighted voting system
which combined a variety of base models. In Section 2 we provide a brief
description of the cDiscount task and data. Section
3 describes our implementations and the results we obtained. Finally, Section 4 concludes with a discussion and the lessons learnt.

%% file: product_classification.tex
\section{e-Commerce Product Classification}
Item categorization is fundamental to many aspects of an
item life cycle for e-commerce sites such as search, recommendation, catalog building etc. It can be formulated as a supervised
classification problem where the categories are the target
classes and the features are the words composing some textual
description of the items. In the context of the cDiscount 2015 challenge\footnote{\url{https://www.datascience.net/fr/challenge/20/details\#tab\_brief55}} the organisers provided the descriptions of e-commerce items and the goal was to develop a system that would perform automatic classification of those products. Table \ref{tbl:exampleInstances} presents some training instances. There were 15,786,886 product descriptions in the training test and 35,066 instances to be classified in the test set. The target categories (classes) were organised in a hierarchy that comprised 3 levels: the top level had 52 nodes, the middle 536 and the lowest level 5,789. The goal of the challenge was to predict the lowest level category for each test instance. It is to be noted that most of the classes were represented in the training set with only a few examples and there were only a few classes with many examples. For instance  40\% of the training instances belong to the 10 most common classes and around 1,500 classes contain from 1 to 30 product items.

\begin{table*}[t]\small\centering
 \begin{tabular}{cllc}
 \toprule
  Categorie3& Description& Libelle& Marque\\
 \midrule
  1000015309& De Collectif aux éditions SOLESMES& Benedictions de l eglise& \\
  1000015309& De Collectif aux éditions SOLESMES& Notice de st benoit lot de 10& \\
  1000010100  & or 750, poids : 3.45gr, diamants : 0.26carats & Bague or et diamants & AUCUNE\\
  1000003407& Champagne Brut - Champagne-Vendu à l'unité-1 x 75cl& Mumm Brut& AUCUNE\\ 
  \bottomrule
 \end{tabular}
 \caption{Part of the training data. We only present the fields ``Description'', ``Libelle'' and ``Marque'' that we used in our implementations. The ``Categorie3'' value was the class to be predicted.}
\label{tbl:exampleInstances}
\end{table*}

%% file: our_approach.tex
\section{Our approach}
We followed a text classification approach working with the textual information provided in the training data. In this context
$x \in \mathbb{R}^d$ represents a document in a vector space  and $y \in \mathcal{Y}=\{1\ldots K\}$ its associated class label where
$|\mathcal{Y}|>2$. The large number of classes in the problem treated as well as the scarcity of data for minority classes
is a typical situation in large-scale systems. For example, we cite here challenges in the same line where in contrast
the available text is much bigger like LSHTC \cite{partalas2015lshtc} and BioASQ \cite{balikas2014results}. 

%There were two different steps in our approach: (i) the pre-processing of the training set which given the instances in the format of Table \ref{tbl:exampleInstances} generates vectors and (ii) the learning algorithm the generates the predictions given the vectors of the pre-processing step. In summary, we found that the pre-processing step (feature selection and feature engineering) was very important and has the potential of boosting our leaderboard score. [Make a figure showing two boxes, one text ``pre-processing'' and another ``cleaning'' and a loop between them.]

\subsection{Setup}

% What to say
% \begin{itemize}
%  \item dataset cleaning and  preparation (stemming, stop-word removel, tokenization, number-text split, strip accents, not-printable symbols, punctuation)
%  \item unigrams, bigrams, trigrams
%  \item a-power
%  \item tf-idf, binary, count values.
%  \item libsvm format
%  \item subsampling (the general idea to overcome the big-volume training set)
%  \item Marque as one-hot-encoding vs as text - missing values
%  \item normalization
%  \item our validation strategy
% \end{itemize}

\paragraph{Data Pre-processing.} Since we used only the textual information that was provided for each training/test instance by the organisers our first task was to clean the data. The pipeline for cleaning the data included: removal of non-ascii characters, removal of non-printable characters, removal of html tags, accents removal, punctuation removal and lower-casing. We also split words that consisted of a text part and a numerical part in to two distinct parts. For instance, ``12cm'' would become ``12'' and ``cm''. In addition, we did not perform stemming, lemmatization and stop-word removal due to the fact that the text spans were small and such operations would result in loss of information \cite{shen2012large}. Finally, we tokenized the remaining text in words using the white space as delimiter.  

\paragraph{Vectorization.} Having cleaned the text, we generated the vectors using an one-hot-encoding approach. We experimented with binary one-hot-encoded vectors (i.e. a word exists or not), with term frequency vectors (how many times each word occurred in the instance) and with the $tf-idf$ (term frequency, inverse document frequency) weighting scheme. In the early stages of the challenge, we found that the latter performed the best in our experiments and we used it exclusively. To calculate
$tf-idf$ vectors we smoothed $idf$ and applied the sublinear $tf$ scaling 1+$\log(tf)$. Finally, each vector has been normalized to a unit vector.

Classification of short documents can benefit from successful feature selection of $n$-grams. The size of the vocabulary of our cleaned dataset combined with the large number of training instances was prohibitively large for using all the $n$-grams with $n=1,2,3$. On the other hand, selecting too many features may lead to over-fitting. As a result, selecting a representative part of those $n$-grams required careful tuning. 

Apart from the ``Description'' and ``Libelle'' fields that we concatenated, we also used the ``Marque'' field. We examined two ways of integrating this information in our pipeline: by concatenating its value with the already existing text of ``Description'' and ``Libelle'' and by generating binary flags for each of the values of the field seen on the training set. Either way, compared to feature generation using only ``Description'' and ``Libelle'' benefited our models. 

After generating the vectors we applied the $\alpha$-power normalization, where each vector $x=(x_1,x_2,\ldots,x_d)$ is
transformed to $x^{power}=(x_1^{\alpha},x_2^{\alpha},\ldots,x_d^{\alpha})$. This normalisation has been used in 
computer vision tasks \cite{Jegou}. The main intuition is that it reduces the effect of the most common
words. After the transformation the vector is normalized
to the unit vector. We found that taking the root ($a=0.5$) of the values of the features consistently benefited the performance. 

%Finally, before feeding the vectors to the learning algorithm, we transformed them to unit vectors which is also known \cite{fan2008liblinear} to improve  performance and accelerate the calculations.

\paragraph{Subsampling} As the training dataset was highly imbalanced the learned models would be biased towards the big classes. For this reason and also
in order to speed up the vectorization process as well as the training of the systems we randomly sampled the data by downsampling the majority classes. 
This procedure helped to improve the  performance of all the single systems (around +2.5\% to our best single system) and also reduced the training time of the base models. The size of the vocabularies for the several sub-samples ranged from around 1 million unigrams for around half of the data to 1.6 millions of unigrams for the full dataset.

\paragraph{Tuning and Validation Strategy}
Tuning the hyper-parameters of our models was an important aspect of our approach. Performing $k$-fold cross validation in 15 million training instances with thousands of features (corresponding to the vocabulary size of the training set) was prohibitive given our cpu resources. We overcame this problem by performing hyper-parameter tuning locally,  in a subsample of the training set. The subsample consisted of the instances of 1500 classes randomly selected. This approach allowed us to accelerate the calculations.   Note that we also validated the applicability of our tuning strategy using the public part of the leaderboard; the decisions that improved the accuracy of our models locally had the same effect on our scores on the public leaderboard. A trick that we found useful was to use the current best submission on the public board as golden standard in order to validate the trained models. This helped us to avoid unnecessary submissions.

\subsection{Training and Prediction}
We relied on linear models as our base learning choice due to their efficiency on high-dimensional tasks like text classification.
Support Vector Machines (SVMs) are well known for achieving state-of-the-art performance in such tasks. For learning the base
models we used the Liblinear library which can support linear models learning in high dimensional datasets \cite{fan2008liblinear}.

We tried two main strategies: a) flat classifiers which ignore the hierarchical structure among the classes  and b) hierarchical 
top-down classification. For flat classification we followed the One-Versus-All (OVA) approach with a complexity $O(K)$ in the number
of classes. For top-down classification we trained a multi-class classifier for each parent in the hierarchy of products and during
prediction we started from the root and selected the best class according to the current multi-class classifiers. We iteratively proceeded
until reaching a leaf node. Note that top-down hierarchical classification has a logarithmic complexity to the number of classes which accelerates the training and prediction processes significantly. 

Trying to explore different learning methods that would also help to diversify the ensemble, we experimented with $k$-Nearest Neighbors classifiers, Rochio classifiers also known as Nearest-Centroid classifiers and online stochastic gradient-descent methods. Although widely studied we found that those methods did not give satisfactory results. For instance, our 3-Nearest Neighbors runs using 200 thousands unigram features, $tf-idf$ feature representation achieved accuracy score 41.62\% in the public leaderboard. We also experimented with text embeddings using the word2vec tool \cite{mikolov2013efficient}; generating text representations in  a low dimensional space (200 dimensions) and 3-NN as a category prediction approach improved over using the $tf-idf$
representation but was still away from performing competitively. For the above mentioned reasons we only report results in the rest of the report for SVMs. In the majority of the models we used $L_2$-regularized $L_2$-loss SVMs in the dual and we set $C=1.0$ with bias ($-B$ $1$).  

\subsection{Ensembling}
Our final solutions were based on averaging the models in the ensemble which contained 103 models. We experimented with simple voting as well as
with weighted voting schemes. Simple voting was always improving performance when it was calculated in a fraction of the whole ensemble containing
only the best performing models. A simple approach was to order the models according to their performance (in terms of accuracy) with respect to the current
best model on the leader board. Then a simple majority voting was applied on around 20\%-30\% of the ordered classifiers. This procedure would create
a homogeneous sub-ensemble that would likely reduce the variance.

Also, we employed a weighted voting scheme weighting more few of our best single models. Our final best submission (64.20\% on the private board)
was a weighting voting ensemble giving bigger weights to the two best models. Weighted voting consistently improved accuracy about 1.2\%-1.8\%. 
\subsection{Software and Hardware}
During the challenge we used scikit-learn \cite{scikit-learn} as well as our own scripts to pre-process the raw data. For training the models we experimented with Liblinear, scikit-learn and Vowpal Wabbit. We had full access to a machine with 4 cores at 2,4Ghz and 16Gb of RAM and limited access to a shared machine with 24 cores at 3,3Ghz and 128 Gb of RAM. In the first machine we ran our experiments with up to 270 thousand features and in the second the experiments with more features that required more memory.

% \begin{itemize}
%  \item SVM, Knn, Rochio
%  \item tf-idf vs w2v
%  \item Hierarchical Vs Flat
%  \item Online gradient descent vs liblinear
%  \item ensembing
% \end{itemize}

%% file: results.tex
\section{Results}
% Description of out best submissions. One with the subsampling, one with the 2M features. 

\begin{table*}[ht]
\centering
 \begin{tabular}{llcc}
 \toprule
   & Description & Public & Private \\
 \midrule
 1 & 200K unigrams & 61.09 & 60.77 \\
 2 & 200K unigrams, $\alpha$=0.5 & 61.60 & 61.25 \\
 3 & 250K unigrams & 61.142 & 60.77\\
 4 & 300K unigrams & 61.148 & 60.87\\
 5 & 250K unigrams, 250K bigrams & 61.79 & 61.37\\
 6 & 200K unigrams, 400K bigrams, $\alpha$=0.5 & 62.09 & 61.76 \\
 7 & 200K unigrams, 400K bigrams, $\alpha$=0.5, ``Marque'' as binary feature & 62.64 & 62.15 \\
 8 & 1,2 M unigrams, bigrams, trigrams, $\alpha$=0.5 & 62.28 & 61.99\\
 9 & 2M unigrams, bigrams, trigrams  & 62.35 & 61.83 \\
 10& 2M unigrams, bigrams, trigrams, $\alpha$=0.5 & 63.30 &  62.99 \\
 \bottomrule
 \end{tabular}
\caption{A subset of our base model submissions with their scores in the public and private leaderboard}\label{tbl:someResults}
\end{table*}

We provide in Table \ref{tbl:someResults} a subset of our submissions along with the public and private scores we obtained. 
Comparing the pairs of submissions (1),(2) and (9), (10) it is clear that the $\alpha$-power transformation helps the performance with respect to accuracy. From submissions (1), (3) and (4) one can see that by increasing the number of the unigram features the performance increases. However, at around 300 thousand features the improvements becomes negligible. The pairs of submissions (3) , (5) and (2), (6) demonstrate the advantage of adding bigrams apart from unigrams to the feature set. Going further and adding, apart from unigrams and bigrams, trigrams also improves the performance as indicated by comparing submissions (8), (9) and (10) with the rest of the Table. Note that in each of the above cases the features were selected with criterion their frequency. For instance in submission (1) we selected the 200k most common unigrams, and in submission (9) the most common features between all unigrams, bigrams and trigrams. 

We tested also several hierarchical models using 200,000 unigrams for each parent node in the hierarchy. The best submission was one with the first level pruned achieving 60.61\% on the private board while the fully hierarchical model got only 58.88\%. Note that in the case of the pruned hierarchy we remove a step during prediction
and thus reduce the propagation errors. Several other hierarchical models were used to increase the variability of the ensemble.

\begin{table*}[!ht]
\centering
 \begin{tabular}{llccl}
 \toprule
   & Description & Public & Private & Coverage \\
 \midrule
 1 & 270K unigrams, $\alpha=0.5$, half data & 63.56 & 63.11 & 3,208 \\
 2 & Weighted voting 1 & 64.55 & 64.20 & 3,128\\
 3 & Weighted voting 2 & 64.57 & 64.14 & 3,116 \\
 \bottomrule
 \end{tabular}
\caption{Our best submissions with their scores in the public and private leaderboard}\label{tbl:bestResults}
\end{table*}

Table \ref{tbl:bestResults} presents the best single system which was trained in approximately half of the data where
the ``Marque'' was concatenated directly in the description.  Also we present the results on the two best
weighted voting systems using a total of 103 base models. Additionally, we report the coverage of each system which shows how many
classes were detected during the prediction phase.

\section{Conclusion and Discussion}
Participating in the cDiscount 2015 was a nice and interesting experience with several lessons learnt. We shortly discuss below the most important ones: 
\begin{itemize}
 \item Data Preparation. Although short text classification for e-Commerce product is not a new task the cDiscount problems had two particularities: the huge number of training instances and the big vocabulary even after carefully cleaning the dataset. We found that feature selection and feature engineering performed by studying the statistics of the dataset, the predictions of base models and by trying to identify patterns for the classes played an important role. As an example we would like to highlight the ``Marque'' field. We found in the late stages of the challenge that by using this information our models could benefit significantly. 
 
 \item Learning tools. From the early stages of the challenge we decided to use SVMs which are known to perform well in such problems. It it the case, however, that the more tools one can use efficiently the better since there is no strategy that works in every setting. Unfortunately, we did not perform an exhaustive hyper-tuning of Vowpal Wabbit on-line learning system which had the potential to provide a high performing system.

 \item Ensemble methods. Base models such as SVMs can obtain satisfactory performance in classification tasks. In the framework of a challenge, however, ensemble methods have the potential of deciding the winner. Ensemble methods (stacking, feature weighted linear stacking etc.) with a pool of strong and diversified models can improve performance by several units of accuracy. Here we relied on weighted voting processes but we believe that using more sophisticated techniques would have helped us.   
\end{itemize}